\documentclass[letterpaper]{article} 
\usepackage{aaai25} 
\pdfoutput=1
\usepackage{times}  
\usepackage{helvet}  
\usepackage{courier}  
\usepackage[hyphens]{url}  
\usepackage{graphicx} 
\usepackage{natbib}  
\usepackage{caption} 
\usepackage{algorithm}
\usepackage{algorithmic}
\usepackage{amsfonts}
\usepackage{booktabs}
\usepackage{comment}
\usepackage[switch]{lineno}
\usepackage{newfloat}
\usepackage{multirow}
\usepackage{amsmath}
\usepackage{listings}
\DeclareCaptionStyle{ruled}{labelfont=normalfont,labelsep=colon,strut=off} 
\floatstyle{ruled}
\newfloat{listing}{tb}{lst}{}
\floatname{listing}{Listing}
\title{IPDN: Image-enhanced Prompt Decoding Network for \\ 3D Referring Expression Segmentation}

\author{
    Qi Chen\equalcontrib\textsuperscript{\rm 1},
    Changli Wu\equalcontrib\textsuperscript{\rm 1},
    Jiayi Ji\thanks{Corresponding author}\textsuperscript{\rm 1 2},
    Yiwei Ma\textsuperscript{\rm 1},
    Danni Yang\textsuperscript{\rm 1},
    Xiaoshuai Sun\textsuperscript{\rm 1}
}
\affiliations{
    \textsuperscript{\rm 1}Key Laboratory of Multimedia Trusted Perception and Efficient Computing, Ministry of Education of China, Xiamen University, 361005, P.R. China.
    
    \textsuperscript{\rm 2}National University of Singapore.

    chenqi@stu.xmu.edu.cn, wuchangli@stu.xmu.edu.cn, jjyxmu@gmail.com, mayiwei@stu.xmu.edu.cn, yangdanni@stu.xmu.edu.cn, xssun@xmu.edu.cn
}

\begin{document}

\maketitle

\begin{abstract}
3D Referring Expression Segmentation (3D-RES) aims to segment point cloud scenes based on a given expression. However, existing 3D-RES approaches face two major challenges: feature ambiguity and intent ambiguity. Feature ambiguity arises from information loss or distortion during point cloud acquisition due to limitations such as lighting and viewpoint. Intent ambiguity refers to the model's equal treatment of all queries during the decoding process, lacking top-down task-specific guidance. In this paper, we introduce an Image-enhanced Prompt Decoding Network (IPDN), which leverages multi-view images and task-driven information to enhance the model's reasoning capabilities. To address feature ambiguity, we propose the Multi-view Semantic Embedding (MSE) module, which injects multi-view 2D image information into the 3D scene and compensates for potential spatial information loss. To tackle intent ambiguity, we designed a Prompt-Aware Decoder (PAD) that guides the decoding process by deriving task-driven signals from the interaction between the expression and visual features. Comprehensive experiments demonstrate that IPDN outperforms the state-of-the-art by 1.9 and 4.2 points in mIoU metrics on the 3D-RES and 3D-GRES tasks, respectively. 
\end{abstract}

%
\begin{links}
     \link{Code}{https://github.com/80chen86/IPDN}
\end{links}

\section{Introduction}
3D Referring Expression Segmentation (3D-RES) presents significant potential applications in areas such as virtual reality, augmented reality, robotics navigation, and human-computer interaction. The goal of this task is to segment the object pointed to by a given textual description from a point cloud scene ~\cite{tgnn,stmn}. 

The earliest approaches ~\cite{tgnn} to 3D-RES employed a two-stage paradigm: first, they used an instance segmentation network to generate proposals, and subsequently matched these proposals with the text to compute matching scores, leading to the final segmentation result. However, this methodology was found lacking in both efficiency and effectiveness ~\cite{stmn}. Consequently, recent studies ~\cite{stmn,refmask3d,segpoint,3dgres} have pivoted toward adopting a one-stage query-based paradigm. For example, 3D-STMN ~\cite{stmn} achieves efficient segmentation by directly matching text with superpoints, while MCLN ~\cite{mcln} and some other works ~\cite{segpoint,unified,xu2024unified} enhance performance by coupling the 3D-RES task with other tasks for joint multi-task training.

\begin{figure}[t]
\centering
\includegraphics[width=0.99\columnwidth]{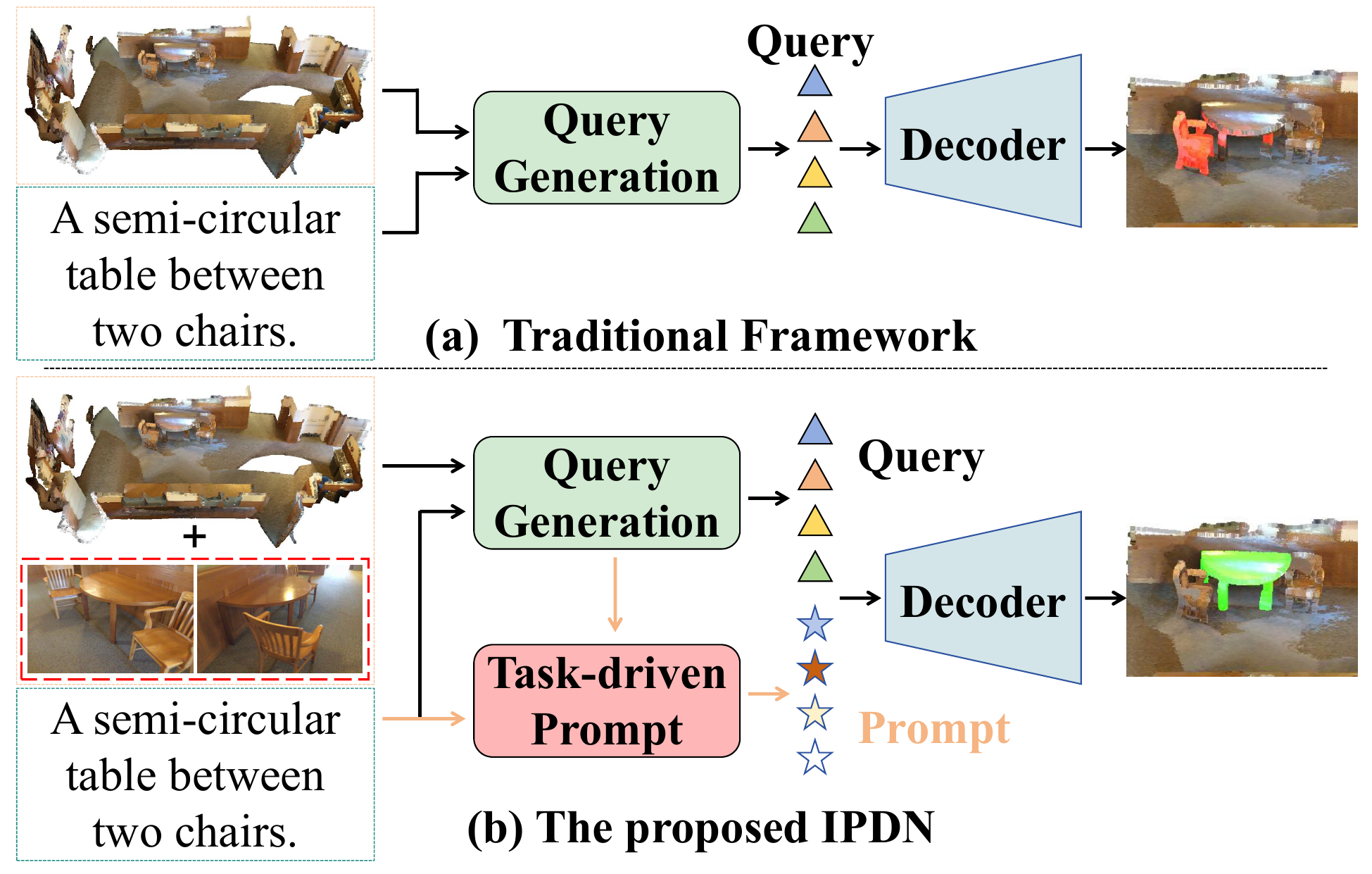} 
\caption{The pipeline of (a) the previous traditional query-based framework and (b) our method.}
\label{fig:1}
\end{figure}

However, despite the promising results these methods have achieved, they still come with certain limitations: (1) \textbf{Feature ambiguity}: Existing approaches rely solely on point cloud data for visual information extraction. However, point cloud data often suffers from information loss due to factors such as lighting, viewing angles, and sampling rates during collection, making it challenging to reproduce real-world scenes faithfully. Consequently, extracting high-quality features exclusively from point cloud data becomes difficult. Compared to 2D data, acquiring and annotating 3D data is far more challenging \cite{2d_help_3d}, limiting the rapid advancements seen in large-scale vision-language pretraining for 2D domains\cite{2d_vlm:1,2d_vlm:2,2d_vlm:3}. Therefore, purely visual 3D backbones ~\cite{backbone:1,backbone:2,backbone:3,backbone:4,backbone:5,backbone:6,backbone:7,backbone:8} struggle to align extracted features with textual representations.
(2) \textbf{Intent ambiguity}: For all queries, they are treated with equal importance, similar to purely visual 3D segmentation~\cite{ins_seg:1,ins_seg:2,ins_seg:3,ins_seg:4,ins_seg:5}. However, in 3D-RES, only the target object described in the text needs to be segmented. Ideally, queries relevant to the text should be prioritized. Yet, current methods ~\cite{stmn,3dgres,refmask3d} do not highlight these relevant queries, leading to the model having to implicitly learn the distinction between relevant and irrelevant queries, significantly increasing the difficulty of the learning process.
 
To address the above issues, we introduce the Image-enhanced Prompt Decoding Network (IPDN), which leverages multi-view images and task-driven information in a top-down approach to unleash the model's reasoning capabilities. 
As shown in Fig.~\ref{fig:1}, to tackle the feature ambiguity issue, we propose the Multi-view Semantic Embedding (MSE) strategy. MSE employs CLIP~\cite{clip} to extract 2D image features, which are then fused with 3D point cloud features to significantly enhance visual representation. Additionally, Spatial-aware Attention is incorporated to address the absence of spatial positional relationships in 2D features. This approach results in visual features with superior representational power, enriched with text prior knowledge from CLIP, facilitating better alignment with textual features. 
To address the intent ambiguity issue, we designed a Prompt-aware Decoder (PAD) that guides the decoding process using task-driven signals. Through the Task-driven Prompt module, we generate prompts that emphasize the relevance of each query to the text, effectively injecting task-specific information into the model and significantly reducing the learning complexity.
Extensive qualitative and quantitative experiments on the ScanRefer~\cite{scanrefer} and Multi3DRefer~\cite{multi3drefer} datasets validate the superior performance of IPDN, surpassing the current state-of-the-art (SOTA) by 1.9 and 4.2 points in mIoU metrics on the 3D-RES and Generalized 3D Referring Expression Segmentation (3D-GRES) tasks, respectively.

To sum up, our main contributions are as follows:

\begin{itemize}
    \item We identify two critical challenges in the 3D-RES task, \emph{i.e.}, feature ambiguity and intent ambiguity, and propose a novel method, IPDN, to effectively address them.
    \item IPDN comprises two essential modules, \emph{i.e.}, MSE and PAD. The MSE integrates multi-view image information into 3D representations while restoring spatial information lost. The PAD pre-processes task-related signals to guide the decoding process with greater precision.
    \item Extensive experiments show that our IPDN outperforms existing state-of-the-art methods, delivering significant improvements in both 3D-RES and 3D-GRES tasks.
\end{itemize}

\section{Related Work}

\begin{figure*}[t]
\centering
\includegraphics[width=0.94\textwidth]{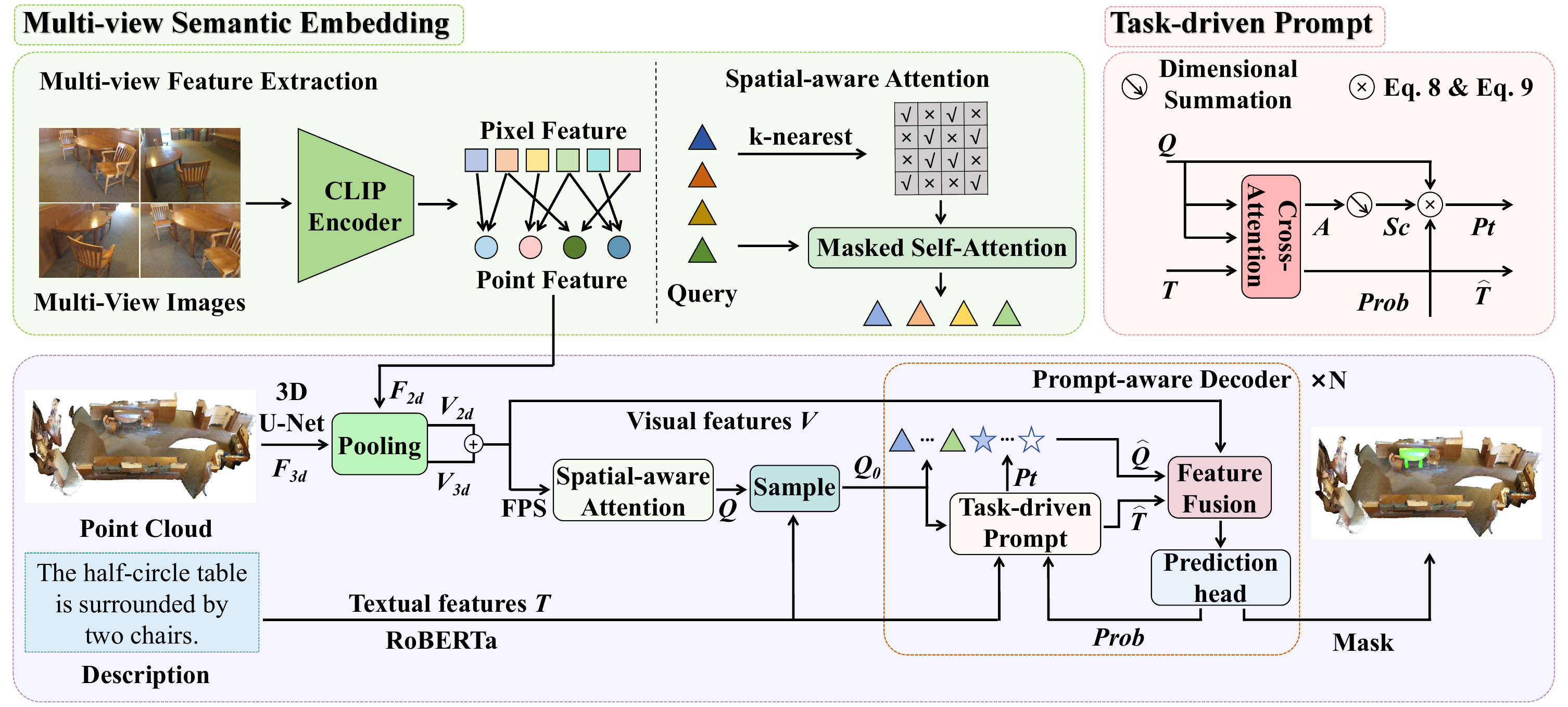}
\caption{The overview of our framework.}
\label{fig:overview}
\end{figure*}

\subsection{3D Referring Expression Comprehension} 3D Referring Expression Comprehension (3D-REC) task is to predict a bounding box for objects indicated by text. Existing approaches to the 3D-REC task can largely be categorized into two types: two-stage ~\cite{3drec_2:1,3drec_2:2,3drec_2:3,3drec_2:4,3drec_2:5,3drec_2:6,3drec_2:7,3drec_2:8,3drec_2:9,3drec_2:10} and one-stage ~\cite{3drec_1:1,3drec_1:2}. Two-stage methods first employ detection models to generate proposals, then use a series of strategies to semantically match the text with these proposals in order to identify the target object. 

\subsection{3D Referring Expression Segmentation} Unlike the relatively mature studies in 3D-REC and 2D-RES ~\cite{2dres:1,2dres:2,2dres:3,2dres:4,2dres:5,2dres:6,2dres:7,2dres:8}, 3D-RES ~\cite{xrefseg3d,refmask3d,segpoint,unified,xu2024unified} is still in its infancy. As the pioneering work in this domain, TGNN ~\cite{tgnn} adopted a two-stage strategy, leveraging Graph Neural Networks for matching candidate instances with textual descriptions. 3D-STMN ~\cite{stmn} harnessed a one-stage method, significantly enhancing both inference speed and performance. Other approaches, such as MCLN ~\cite{mcln}, capitalized on the similarity between 3D-RES and 3D-REC tasks to facilitate multitask joint learning.

To liberate the 3D-RES task from its constraint of having one and only one target object per sentence, the 3D-GRES task was introduced ~\cite{3dgres}. A distinctive feature of 3D-GRES is that the object referenced by the text may not exist or could be multiple objects, no longer restricted to a single object. 

\subsection{Prompt Learning} Prompt learning generally refers to the augmentation of models with specific prompt information. These prompts can be hand-crafted or automatically learned during the training process. Initially applied in the field of Natural Language Processing (NLP) ~\cite{pl_l:1,pl_l:2,pl_l:3}, prompt learning has since been adapted for use in visual ~\cite{pl_v:1,pl_v:2,pl_v:3} and vision-language ~\cite{pl_vl:1,pl_vl:2,pl_vl:3} models as well. In our model, we utilize a set of prompts generated under textual guidance to instruct the model in differentiating between relevant queries and irrelevant ones.

\section{Method}

In this section, we first introduce the inputs to the decoder, namely how visual features, textual features, and queries are obtained (Sec.~\ref{fe}). Secondly, we detail the Multi-view Semantic Embedding (MSE) strategy (Sec.~\ref{mse}). Then we describe the Prompt-aware Decoder (Sec.~\ref{dec}). Finally, we outline the loss function for the entire model (Sec.~\ref{loss}). An overview of our framework is shown in Fig.~\ref{fig:overview}.

\subsection{Feature Extraction}\label{fe}
\subsubsection{Textual Feature} Given a textual description for the target object, we utilize a pre-trained RoBERTa ~\cite{roberta} to extract word-level embeddings $E\in \mathbb{R}^{N_{t}\times C_{t}}$, where $N_{t}$ denotes the number of tokens, and $C_{t}$ indicates the $C_{t}$-dimensionality of each embedding. In order to have a unified feature dimension $d$ in the decoder, we transform $E$ into textual features $T\in \mathbb{R}^{N_{t}\times d}$ via a linear projection:
\begin{equation}
T=EW_{t},
\end{equation}
where $W_{t}\in \mathbb{R}^{C_{t}\times d}$ are learnable parameters.

\subsubsection{Visual Feature} Given a point cloud scene $P\in \mathbb{R}^{N_{p}\times(3+f)}$, where $N_{p}$ denotes the number of points. Each point carries a 3D coordinate as well as an auxiliary feature vector of $f$ dimensions, such as RGB values and normal vectors. We first employ a Sparse 3D U-Net ~\cite{backbone:3} to extract point-wise features $F_{3d}\in \mathbb{R}^{N_{p}\times C_{p}}$, where $C_{p}$ represents the feature dimensionality. Subsequently, following the approach of ~\cite{ins_seg:2}, we generate $N_{s}$ superpoints $\{SP_{i}\}_{i=1}^{N_{s}}$ ~\cite{superpoint} from the original point cloud and perform superpoint pooling on $F_{3d}$ to obtain 3D superpoint features $S_{3d}\in \mathbb{R}^{N_{s}\times C_{p}}$. Then, a multi-layer perceptron (MLP) is utilized to transform the dimensionality to $d$, yielding the 3D visual features $V_{3d}\in \mathbb{R}^{N_{s}\times d}$:
\begin{equation}
V_{3d}=\text{MLP}(\text{SPPool}(F_{3d})),
\label{spp}
\end{equation}
where $\text{MLP}(\cdot)$ is a learnable multi-layer perceptron, and $\text{SPPool}(\cdot)$ is superpoint pooling operation. The final visual feature $V\in \mathbb{R}^{N_{s}\times d}$ is obtained by the sum of $V_{3d}$ and $V_{2d}$ (introduced in Sec.~\ref{mse}).

\subsubsection{Sparse Query Generation} 
After obtaining the visual features $V$ and textual features $T$, our next step is to utilize both to generate the queries that will be used in the decoder. Specifically, We first perform farthest point sampling ~\cite{fps} on the superpoints (correspond one-to-one with the visual features), followed by spatial-aware attention (introduced in Sec.~\ref{mse}), then resample the results using the sampling module from MDIN ~\cite{3dgres}, and generate the queries through an MLP:
\begin{equation}
    Q_{seed}=V[\text{FPS}(p_{sp})],
\end{equation}
\begin{equation}
    Q_{0}=\text{MLP}(\text{Sample}(\text{SPA}(Q_{seed}), T)),
\end{equation}
where $\text{FPS}(\cdot),\text{Sample}(\cdot)$ and $\text{SPA}(\cdot)$ denote the Farthest Point Sampling algorithm, the sampling module in MDIN and the spatial-aware attention respectively, $[\cdot]$ denotes accessing elements by the index within it, $p_{sp}\in \mathbb{R}^{N_{s}\times 3}$ represents the coordinates of the superpoints, $Q_{seed}\in \mathbb{R}^{2m\times d}$ is the seed query, and $Q_{0}\in \mathbb{R}^{m\times d}$ ($m<<N_{s}$) is initial query.

\subsection{Multi-view Semantic Embedding}\label{mse}

\subsubsection{Multi-view Feature Extraction} 

The 3D features extracted solely from point cloud data are limited in representational capacity due to the information loss of point clouds and insufficient alignment with the language modality. To address this issue, we propose a Multi-View Semantic Embedding (MSE) strategy. This approach enhances the visual features by extracting well-aligned multi-view semantics and injecting them back into the original 3D features through 2D-3D projection.

Specifically, given $N_{I}$ images $\{I_{i}\}_{i=1}^{N_{I}}$ of the point cloud scene from different perspectives, we first extract patch-level 2D features using the CLIP ~\cite{clip} visual encoder, which is pre-aligned with visual-language tasks. To accommodate camera parameters, we upsample these features to the original image resolution via interpolation, resulting in pixel-level 2D features $\{F_{i}^{img} \in \mathbb{R}^{H \times W \times C_{I}}\}_{i=1}^{N_{I}}$, where $C_{I}$ denotes the feature dimension, and $H$ and $W$ represent the height and width of the image, respectively. Next, we project the 2D pixel coordinates into the 3D point cloud space using the camera parameters. Similar to previous works ~\cite{proj:1,proj:2,proj:3,proj:4}, for a pixel coordinate $(u, v)$, given the intrinsic camera parameters $\mathcal{K} \in \mathbb{R}^{3 \times 3}$, extrinsic parameters $\mathcal{R} \in \mathbb{R}^{3 \times 3}$ and $\mathcal{T} \in \mathbb{R}^{3 \times 1}$, and depth $\mathcal{D} \in \mathbb{R}$, we obtain the corresponding 3D coordinates $(x, y, z)$ through 2D-3D projection:
\begin{equation}
\left[\begin{matrix}x\\y\\z\end{matrix}\right]=\text{Project}(u,v)=\mathcal{R}(\mathcal{K}^{-1}\left[\begin{matrix}u\\v\\1\end{matrix}\right]\cdot\mathcal{D})+\mathcal{T}.
\end{equation}

After the projection, all 2D pixel features are assigned 3D coordinates $p_{3d} \in \mathbb{R}^{HWN_{I} \times 3}$. To inject these multi-view features into the point cloud, we apply spherical querying to $p_{3d}$ in the point cloud scene. This technique assigns each pixel feature to the points within a sphere centered at its 3D coordinate, thus embedding multi-view semantic information. For points residing in multiple spheres, the final multi-view feature is computed as the average of the pixel features associated with that point. In this way, we obtain the multi-view semantic features $F_{2d} \in \mathbb{R}^{N_{p} \times C_{I}}$ for all points, which are then processed similarly to $F_{3d}$ (Eq.~\ref{spp}) to derive the 2D visual features $V_{2d}\in \mathbb{R}^{N_{s}\times d}$. Finally, we get the visual feature $V\in \mathbb{R}^{N_{s}\times d}$ by summing the $V_{3d}$ and $V_{2d}$.

\subsubsection{Spatial-aware Attention}\label{spa} 

While incorporating multi-view semantics improves visual representation and visual-language alignment, it also introduces limitations inherent to 2D images, such as the absence of spatial positional information and potential multi-view conflicts. Specifically, each image has a restricted field of view and lacks depth information, complicating the determination of 3D object positions and inter-object distances. To mitigate these issues, we use a spatial-aware attention mechanism to incorporate explicit 3D spatial relationships, enhancing spatial positioning. Additionally, due to the high computational cost and inefficiency of operating directly at the superpoint level, we implement efficient spatial-aware attention on the sparse seed query $Q_{seed}$, which is more manageable on our GPU.

First, we construct a $k$-nearest neighbor matrix $M \in \mathbb{R}^{2m \times 2m}$, where the element $M_{ij}$ in the $i^{th}$ row and $j^{th}$ column indicates whether the $j^{th}$ query is among the $k$ nearest queries to the $i^{th}$ query. If the $j^{th}$ query is within the $k$ nearest neighbors of the $i^{th}$ query, $M_{ij}$ is set to True; otherwise, it is set to False. The coordinates of the queries are obtained from the corresponding superpoint coordinates. Then, we use $M$ as a mask to perform self-attention on the seed queries $Q_{seed}$, producing the output $Q \in \mathbb{R}^{2m \times d}$ as the input to the sample module:
\begin{equation}
Q=\text{SPA}(Q_{seed})=\text{Masked\_Self}(Q_{seed},M),
\end{equation}
where $\text{Masked\_Self}(\cdot)$ denotes the masked self-attention.

\subsection{Prompt-aware Decoder}\label{dec}

Previous query-based methods~\cite{stmn, refmask3d, mcln} inherit the instance segmentation approach~\cite{ins_seg:2,ins_seg:3,ins_seg:4} to handling queries, which does not distinguish the importance of different queries. However, this approach is not well-suited for the 3D-RES task, which aims to segment objects indicated by text rather than all objects. This means that queries related to the text should be prioritized. To help the model better differentiate the importance of queries and reduce the learning difficulty, we introduce task-driven prompt learning in the decoder. By dynamically generating a set of text-relevant prompts, these prompts guide the model during the decoding process to identify which queries are more important and more likely to correspond to the target object.

\subsubsection{Task-driven Prompt} 

To design reliable prompts tailored for 3D-RES task, we first measure the relevance between the text and queries using cross-attention scores. Specifically, we perform a cross-attention operation by using the text features $T$ as the query and the Sparse Queries $Q_{l}$ from the $l^{th}$ layer as the keys and values within the attention mechanism:
\begin{equation}
    \hat{T}_{l},A_{l}=\text{Cross}(T,Q_{l},Q_{l}),
\end{equation}
where $\hat{T}_{l}\in \mathbb{R}^{N_{t}\times d}$, $A_{l}\in \mathbb{R}^{N_{t}\times m}$, and $Q_{l}\in \mathbb{R}^{m\times d}$ denote the text features, attention scores, and sparse queries at the $l^{th}$ layer, respectively. $\text{Cross}(\cdot)$ denotes the cross-attention operation ~\cite{attention}. Then, by summing the attention scores $A_{l}$ across the first dimension, we initially obtain the relevance scores $Sc_{l}\in \mathbb{R}^{m}$ indicating how closely each query is associated with the given textual description.

After obtaining the scores $Sc_{l}$, a intuitive approach would be to directly apply the Softmax function to determine the desired relevance. However, most queries are irrelevant to the text description, and their scores essentially act as noise, which should be minimized. To address this, we introduce a threshold filtering operation to filter out irrelevant queries as much as possible, making the prompts more reliable.

Specifically, we utilize the probability $Prob_{l-1} \in \mathbb{R}^{m}$, generated by the prediction head of the upper layer queries, to filter out queries. This probability represents the likelihood that a query corresponds to the target instance. For queries with probabilities below the threshold $r$, their relevance scores are set to negative infinity, meaning their values will be 0 after applying the Softmax function. Finally, the Softmax function is applied to the relevance scores, and the results are multiplied by the queries, producing prompts that guide the model in distinguishing between relevant and irrelevant queries. The process can be formulated as follows:
\begin{equation}
\hat{Sc}_{l}^{j}=\left\{
    \begin{aligned}
    -\infty  ,&\ \ \ \ Prob_{l-1}^{j}<r \\ 
    Sc_{l}^{j} ,&\ \ \ \ Prob_{l-1}^{j}\geq r
    \end{aligned}
\right.,
\end{equation}
\begin{equation}
Pt_{l}=Q_{l}\cdot \text{Softmax}(\hat{Sc}_{l}),
\end{equation}
\begin{equation}
\hat{Q}_{l}=\text{Concat}(Q_{l},Pt_{l}),
\end{equation}
where $r$ is a hyperparameter, $j$ denotes the $j^{th}$ element, $Pt_{l}\in \mathbb{R}^{m\times d}$ represents the prompts, $\text{Concat}(\cdot)$ denotes the Concatenation operation, $\hat{Q}_{l}\in \mathbb{R}^{2m\times d}$ stands for the queries with the prompts attached, and all subscripts $l$ indicate the $l^{th}$ layer.

\subsubsection{Feature Fusion \& Prediction Head} We follow the feature fusion method outlined in MDIN ~\cite{3dgres}, utilizing the query $\hat{Q}_{l}$ to integrate the textual features $\hat{T}_{l}$ and visual features $V$, thereby updating the queries under the guidance of the prompts. The specific formula is presented as follows:
\begin{equation}
\begin{aligned}
\bar{Q}_{l}=\text{Abandon} & (\text{Cross}(\hat{Q}_{l},\hat{T}_{l},\hat{T}_{l})\\
 & +\text{Self}(\hat{Q}_{l}) \\
 & +\text{Cross}(\hat{Q}_{l},V,V)),
\end{aligned}
\end{equation}
where $\text{Self}(\cdot)$ denotes the self-attention operation, $\text{Abandon}(\cdot)$ denotes the discarding of the prompts, and $\bar{Q}_{l}\in \mathbb{R}^{m\times d}$ represents the updated queries. Considering the difference in scale between the queries and superpoints, we apply Spatial-aware Attention once again at the end of each layer for feature enhancement and to generate the query $Q_{l+1}$ for the next layer: 
\begin{equation}
    Q_{l+1}=\text{SPA}(\bar{Q}_{l}).
\end{equation}

Before going to the next layer, $Q_{l+1}$ will pass through a prediction head to generate $Mask_{l}$ and $Prob_{l}$ in $l^{th}$ layer:
\begin{equation}
    Mask_{l}=Q_{l+1}(VW_{mask})^{T}, 
\end{equation}
\begin{equation}
    Prob_{l}=Q_{l+1}W_{prob},
\end{equation}
where $W_{mask}\in \mathbb{R}^{d\times d}$ and $W_{prob}\in \mathbb{R}^{d\times 1}$ are learnable parameters, superscript $T$ indicates matrix transpose, $Mask_{l}\in \mathbb{R}^{m\times N_{s}}$ represents the predicted masks for every query and $Prob_{l}\in \mathbb{R}^{m}$ represents the likelihood that a query corresponds to the target instance.

Following MDIN~\cite{3dgres}, we select the query with the highest $Prob$ value and binarize its corresponding mask to generate the prediction during inference in the 3D-RES task. In the 3D-GRES task, we merge the binary masks of all queries with $Prob$ values greater than 0.5 to produce the final prediction.

\subsection{Loss}\label{loss}
The loss of our method primarily consists of three components. The first component is the basic loss $\mathcal{L}_{b}$, which is applied only on the queries corresponding to the target instance ~\cite{3dgres}:
\begin{equation}
\mathcal{L}_{b}=\text{BCE}(M^{+},M^{tgt})+\text{DICE}(M^{+},M^{tgt}),
\end{equation}
where $M^{+}$ represents the mask output by the prediction head for the query corresponding to the target instance, $M^{tgt}$ 
is the ground truth mask for the target instance, $\text{BCE}(\cdot)$ denotes the Binary Cross-Entropy loss, and $\text{DICE}(\cdot)$ refers to the Dice loss ~\cite{dice}.

The second part is the probability loss $\mathcal{L}_{p}$, which is used to supervise $Prob$, that is, the probability associated with the query corresponding to the target instance:
\begin{equation}
\mathcal{L}_{p}=\text{BCE}(Prob,L^{tgt}),
\end{equation}
where the label $L^{tgt}\in \{0,1\}^{m}$ indicates whether the query corresponds to the target instance, with 1 representing a positive match and 0 representing a negative match, and $Prob$ is the probability output by the prediction head.

The third part is the contrastive learning loss $\mathcal{L}_{c}$, which is used to align the text features with their corresponding queries. Here, we adopt the approach used in EDA ~\cite{eda}. 

The final loss $\mathcal{L}$ is calculated as the weighted sum of $\mathcal{L}_{b}$, $\mathcal{L}_{p}$ and $\mathcal{L}_{c}$:
\begin{equation}
\mathcal{L}=\lambda_{b}\mathcal{L}_{b}+\lambda_{p}\mathcal{L}_{p}+\lambda_{c}\mathcal{L}_{c},
\end{equation}
where $\lambda_{b}$, $\lambda_{p}$ and $\lambda_{c}$ are hyperparameters.

\begin{table*}
    \centering
    \setlength{\tabcolsep}{1mm}{
        \begin{tabular}{l|cccc|cccc|c}
            \toprule 
            & \multicolumn{4}{c|}{Acc@0.25} 
            & \multicolumn{4}{c|}{Acc@0.5} & \\
            \multirow{-2}{*}{Method}
            & ZT & ST & MT & All
            & ZT & ST & MT & All
            & \multirow{-2}{*}{mIoU}  \\
            \midrule

            ReLA~{\footnotesize\cite{2dres:6}} 
            & {36.2} / {72.7} & {48.3} / {83.4} & {73.0} & {61.8}
            & {36.2} / {72.7} & {20.4} / {65.5} & {42.4} & {37.4}
            & {42.8} \\

            M3DRef-CLIP~{\footnotesize \cite{multi3drefer}} 
            & {39.2} / {81.6} & {50.8} / {77.5} & {66.8} & {55.7}
            & {39.2} / {81.6} & {29.4} / {67.4} & {41.0} & {37.5}
            & {37.4} \\

            3D-STMN~{\footnotesize\cite{stmn}} 
            & {42.6} / {76.2} & {49.0} / {77.8} & {68.8} & {60.4}
            & {42.6} / {76.2} & {24.6} / {69.2} & {43.9} & {40.9}
            & {43.0} \\

            MDIN~{\footnotesize\cite{3dgres}}
            & \textbf{47.9} / {78.8} & {55.5} / {84.4} & {76.3} & {67.0}
            & \textbf{47.9} / {78.8} & {29.5} / {71.7} & {46.8} & {44.7}
            & {47.5} \\ 

            
            IPDN (Ours) 
            & {39.4} / \textbf{84.1} & \textbf{61.5}  
            / \textbf{88.9} & \textbf{79.6} & \textbf{71.5} 
            & {39.4} / \textbf{84.1} & \textbf{34.7} 
            / \textbf{79.5} & \textbf{52.1} & \textbf{50.0}
            & \textbf{51.7} \\

            \bottomrule
        \end{tabular}
    }

    \caption{The 3D-GRES results on Multi3DRefer. ZT, ST, and MT represent zero target, single target, and multiple targets, respectively. The left and right sides of the ``/" represent the situations with and without distractor objects, respectively. 
    }
    \label{tab:multi3drefer}
\end{table*}

\section{Expriments}

\subsection{Implementation details} In our experiments, we apply the PolyRL strategy to adjust the learning rate starting from 0.0001, with a decay power of 4.0. The batch size is set to 16. The number of queries $m$ is set to 128. The decoder consists of 6 layers. The hyperparameter $k$ in sec.\ref{spa} is set to 8, and the hyperparameter $r$ in sec.~\ref{dec} is 0.75. In the loss function, the weights $\lambda_{b}$, $\lambda_{p}$, and $\lambda_{c}$ are set to 1.0, 0.1, and 0.1 respectively. All experiments are conducted using the PyTorch framework on an NVIDIA GeForce RTX 3090 GPU.

\subsection{Dataset and Evaluation Metrics} 

\subsubsection{ScanRefer} We utilize the ScanRefer dataset~\cite{scanrefer} to evaluate our method, which consists of 51,583 natural language expressions, encompassing 11,046 objects across 800 ScanNet~\cite{scannet} scenes. The evaluation metrics include mean Intersection over Union (mIoU), Acc@0.25, and Acc@0.5.

\subsubsection{Multi3DRefer} We use the Multi3DRefer~\cite{multi3drefer} dataset to evaluate our model's performance on the 3D-GRES task, which differs from 3D-RES in that the number of targets referenced by the text can be arbitrary. The dataset consists of a total of 61926 language descriptions, of which 51583 are directly obtained from ScanRefer. Among these, 6688 descriptions match zero targets, 13178 match multiple targets, and the rest match a single target. The evaluation metric is the same as that used in ScanRefer. When the text refers to zero object, the sample's mIoU is 1 if the model correctly identifies this, otherwise, it is 0.

\subsection{Quantitative Comparison}

\begin{table*}
    \centering
    {
        \begin{tabular}{l|c|ccc|ccc|ccc}
            \toprule &
            & \multicolumn{3}{c|}{Unique ($\sim$19\%)} 
            & \multicolumn{3}{c|}{Multiple ($\sim$81\%)} 
            & \multicolumn{3}{c}{\textbf{Overall}}\\
            \multirow{-2}{*}{Method} & \multirow{-2}{*}{Venue}
            & 0.25 & 0.5 & mIoU 
            & 0.25 & 0.5 & mIoU 
            & \textbf{0.25} & \textbf{0.5} & \textbf{mIoU} \\
            \midrule
            
            TGNN\dag ~\cite{tgnn} & AAAI2021
            & {69.3} & {57.8} & {50.7} 
            & {31.2} & {26.6} & {23.6} 
            & {38.6} & {32.7} & {28.8} \\   

            InstanceRefer\dag ~\cite{3drec_2:4} & ICCV2021
            & 81.6 & 72.2 & 60.4 
            & 29.4 & 23.5 & 21.5 
            & 40.2 & 33.5 & 30.6  \\

            3DRefTR ~\cite{unified} & Arxiv
            & {89.6} &{77.0} & - 
            & {52.3} & {43.7} & - 
            & {57.9} & {48.7} & {41.2} \\

            X-RefSeg3D ~\cite{xrefseg3d} & AAAI2024
            & - & - & - 
            & - & - & - 
            & {40.3} & {33.8} & {29.9} \\

            3D-STMN ~\cite{stmn} & AAAI2024
            & {89.3} & {84.0} & {74.5} 
            & {46.2} & {29.2} & {31.1} 
            & {54.6} & {39.8} & {39.5}  \\
            
            Reanson3D ~\cite{reason3d} & Arxiv
            & {88.4} & {84.2} & {74.6} 
            & {50.5} & {31.7} & {34.1} 
            & {57.9} & {41.9} & {42.0} \\

            SegPoint ~\cite{segpoint} & ECCV2024
            & - & - & - 
            & - & - & - 
            & - & - & {41.7} \\
            
            MCLN ~\cite{mcln} & ECCV2024
            & {89.6} & {78.2} & - 
            & \textbf{53.3} & {45.9} & - 
            & {58.7} & {50.7} & {44.7}  \\
            
            RefMask3D ~\cite{refmask3d} & ACMMM2024
            & {89.6} & {84.7} & - 
            & {48.1} & {40.8} & - 
            & {55.9} & {49.2} & {44.9} \\
            
            MDIN ~\cite{3dgres} & ACMMM2024
            & {91.0} & {87.2} & {76.7} 
            & {50.1} & {44.9} & {41.4} 
            & {58.0} & {53.1} & {48.3}  \\


            IPDN (Ours) & -
            & \textbf{91.5} & \textbf{88.0} & \textbf{77.9} 
            & {53.1} & \textbf{47.0} & \textbf{43.6} 
            & \textbf{60.6} & \textbf{54.9} & \textbf{50.2} \\

            \bottomrule
        \end{tabular}
    }

    \caption{The 3D-RES results on ScanRefer. \dag~The mIoU and accuracy are reevaluated on our machine.}
    \label{tab:scanrefer_benchmark}
\end{table*}

As shown in Tab.~\ref{tab:multi3drefer}, our model significantly outperforms the existing SOTA methods on the 3D-GRES task, achieving an improvement of 4.2 points in mIoU and even 5.3 points in Acc@0.5. It can be observed that, apart from the zero target scenario with distractors where our model performs below, it significantly surpasses the MDIN ~\cite{3dgres} in all other cases. Especially in single-target scenarios, without distractors, our model outperforms the MDIN by nearly eight points on the Acc@0.5 metric. This demonstrates that our task-driven prompt effectively guides the model to focus on more significant queries, thereby enabling more accurate localization of key targets.

We also conducted experiments on the traditional 3D-RES task, as shown in Table~\ref{tab:scanrefer_benchmark}. On the ScanRefer dataset, our proposed IPDN model achieved state-of-the-art performance overall. Specifically, our model outperformed the previous best model, MDIN~\cite{3dgres}, by 2.6, 1.8, and 1.9 points in terms of Acc@0.25, Acc@0.5, and mIoU, respectively. Notably, we observed more significant improvements in challenging scenes with multiple distracting objects. This indicates that our model benefits from more robust multi-view semantic integration and the task-driven prompt, which effectively guides the model to focus on more critical information, thereby enhancing its discriminative ability to accurately identify the target object among multiple instances of the same category.

Thanks to the integration of well-established large-scale pre-trained models from the 2D domain within the MSE module, the visual representations in our model are more robust, enabling it to perform reliably even on rarely seen classes in the training set. To validate this, inspired by \cite{scannet200,ov:1,ov:2,ov:3}, We categorized object classes based on their frequency of appearance in the training set and conducted testing accordingly, as shown in Tab.~\ref{tab:frequency}. Specifically, we categorized all target classes in ScanRefer into three groups. The first group, labeled ``High'', consists of classes that make up more than 1\% of the training set, accounting for approximately 75\% of the total samples. The second group, labeled ``Mid'', includes classes that comprise less than 1\% but more than 0.1\% of the training set, representing about 20\%. The remaining classes, labeled ``Low", make up less than 0.1\% of the training set and account for about 5\% of the samples.

As shown, the performance of 3D-STMN~\cite{stmn} and MDIN~\cite{3dgres} significantly drops for the ``Low'' frequency categories, decreasing by 17.5 and 14.9 points, respectively, compared to the ``High'' group. In contrast, our model shows a decrease of only 6.3 points. When comparing across models, our model outperforms MDIN by more than 10 points in the ``Low'' group. This substantial improvement highlights the enhanced robustness of our model, attributed to the multi-view semantic integration, enabling it to handle infrequent, long-tail samples effectively.

\begin{table}
    \centering
        \begin{tabular}{c|cccc}
            \toprule
            Method & High & Mid & Low & Overall\\
            \midrule

            3D-STMN 
            & {39.1} & {46.7} & {21.6} & {39.5} \\ 

            MDIN 
            & {46.9} & {58.2} & {32.0} & {48.3} \\

            IPDN (Ours)  
            & \textbf{48.5} & \textbf{60.5} & \textbf{42.2} & \textbf{50.2}\\
            
            \bottomrule
        \end{tabular}

    \caption{Test results of the subsets of ScanRefer, divided by frequency, with the mIoU metric. High, Mid, and Low refer to categories that account for more than 1\%, between 0.1\% and 1\%, and less than 0.1\% of the training set, respectively.}
    \label{tab:frequency}
\end{table}

\subsection{Ablation Study}
All of our ablation experiments were conducted on ScanRefer dataset ~\cite{scanrefer}.

\subsubsection{Component Ablation} In our proposed IPDN, the main components include MSE and PAD. To assess the impact of these two components, we conducted an ablation study, as shown in Tab.~\ref{tab:compont_ablation}. The results indicate that omitting both components results in a 2.1-point decrease in mIoU. Introducing MSE improves mIoU by 1.1 points, demonstrating its effectiveness in enhancing visual features. Further inclusion of PAD leads to an additional 1.0-point increase in mIoU, indicating that task-driven prompts effectively guide the model to focus on more important queries, thereby improving segmentation accuracy.

\begin{table}
    \centering
        \begin{tabular}{cc|ccc}
            \toprule
             \textbf{MSE} & \textbf{PAD}
            & \textbf{Acc@0.25} & \textbf{Acc@0.5} & \textbf{mIoU} \\
            \midrule

            $\times$ & $\times$  
            & {58.2} & {52.9} & {48.1}  \\

            \checkmark & $\times$  
            & {58.9} & {53.7} & {49.2}  \\

            \checkmark & \checkmark  
            & \textbf{60.6} & \textbf{54.9} & \textbf{50.2}  \\

            \bottomrule
        \end{tabular}

    \caption{Ablation study on the proposed components. Not using PAD signifies removing the prompt from the decoder.}
    \label{tab:compont_ablation}
\end{table}

\subsubsection{Spatial-aware Attention Ablation} We conducted an ablation study on the hyperparameter $k$ in the Spatial-aware Attention, and the results are shown in Tab.~\ref{tab:k_ablation}. From row 1 and row 2, we can see that when $k$ is small, the expected performance is not achieved. This is because a smaller $k$ represents a smaller receptive field for objects, which may be even smaller than the field of view of the image itself, thus leading to minimal improvement. At the same time, from row 3$~\sim$5, it is evident that a larger $k$ is not always better. When $k$ is too large, objects may attend to distant irrelevant objects, causing a negative effect. In summary, setting $k$ to 8 is considered reasonable.

\begin{table}
    \centering
        \begin{tabular}{cc|ccc}
            \toprule
                & $k$ 
            & \textbf{Acc@0.25} & \textbf{Acc@0.5} & \textbf{mIoU} \\
            \midrule

            1 & 2   
            & {59.1} & {53.9} & {49.3}  \\

            2 & 4   
            & {59.2} & {54.1} & {49.4}  \\

            3 & 6   
            & {59.9} & {54.5} & {49.7}  \\

            4 & 8   
            & \textbf{60.6} & \textbf{54.9} & \textbf{50.2}  \\
            
            5 & 10   
            & {60.5} & {54.6} & {50.0}  \\
            
            \bottomrule
        \end{tabular}

    \caption{Ablation study on the hyperparameter $k$.}
    \label{tab:k_ablation}
\end{table}

\subsubsection{PAD Ablation} We conducted an ablation study on the hyperparameter $r$ in the PAD, and the results are shown in Tab.~\ref{tab:r_ablation}. From the row 1, we can see that when $r$ is 0 (no filtering), there is a significant amount of noise in the prompts, leading to minimal effect, with only a 0.2 mIoU improvement (compared to row 2 in Tab.~\ref{tab:compont_ablation}). From row 2 to row 4, we observe that as $r$ increases, more irrelevant queries are filtered out, resulting in improved prompting effects. From the row 5, we learn that $r$ is not necessarily better when it is larger, because when $r$ is too large, some relevant queries are also filtered out, leading to a decrease in the prompting effect. In summary, 0.75 is a suitable threshold, effectively filtering out irrelevant queries while retaining relevant ones.

\begin{table}
    \centering
        \begin{tabular}{cc|ccc}
            \toprule
                & $r$ 
            & \textbf{Acc@0.25} & \textbf{Acc@0.5} & \textbf{mIoU} \\
            \midrule

            1 & 0   
            & {59.3} & {54.3} & {49.4}  \\

            2 & 0.25   
            & {59.8} & {54.2} & {49.6}  \\

            3 & 0.5   
            & {59.7} & {54.5} & {49.6}  \\

            4 & 0.75   
            & \textbf{60.6} & \textbf{54.9} & \textbf{50.2}  \\
            
            5 & 0.9   
            & {60.3} & {54.4} & {50.0}  \\
            
            \bottomrule
        \end{tabular}

    \caption{Ablation study on the hyperparameter $r$.}
    \label{tab:r_ablation}
\end{table}

\begin{figure}[t]
\includegraphics[width=0.99\columnwidth]{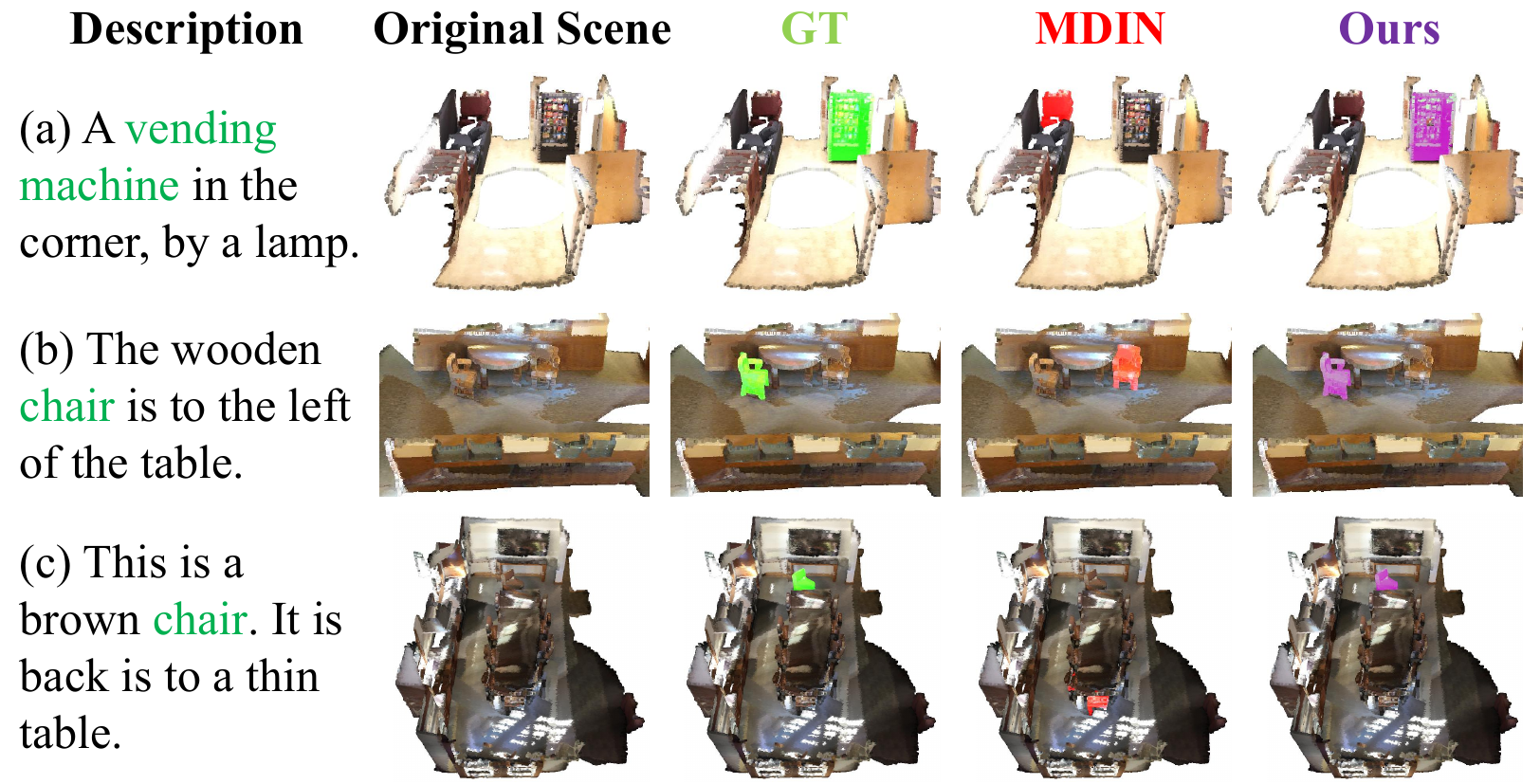} 
\caption{Qualitative comparison between MDIN and ours.}
\label{fig:visual}
\end{figure}

\subsection{Qualitative Comparison} In this section, we visualized a set of representative examples in ScanRefer dataset, as shown in Fig.~\ref{fig:visual}. It can be seen from the figure that our model demonstrates stronger reasoning capabilities compared to MDIN ~\cite{3dgres}. Specifically, in case (a), there is no distracting object present in the scene, only a vending machine, but MDIN still fails to identify it. This is because, within the ScanRefer training dataset of over 30,000 samples, there are only 10 samples where the target is a vending machine, which is insufficient for the model to recognize such an object. However, models with large-scale 2D pre-training do not suffer from this issue and can well identify vending machines, allowing our model to accurately locate the target. In case (b), the concept of ``left" is involved, which is perspective-dependent. Since three-dimensional space theoretically contains an infinite number of perspectives, purely 3D models have difficulty distinguishing left from right. In contrast, the perspective in 2D images is fixed, which provides significant assistance in handling such cases. Finally, in case (c), thanks to the powerful prompting ability of our task-driven prompts, even when there are nearly ten distractor objects present, our model can still accurately locate the target object.

\section{Conclusion}

In this paper, we focus on addressing feature ambiguity and intent ambiguity by introducing the Image-enhanced Prompt Decoding Network (IPDN). To overcome feature ambiguity, we propose the Multi-view Semantic Embedding (MSE) module, which incorporates multi-view 2D image information into the 3D scene, compensating for any potential spatial information loss. To resolve intent ambiguity, we developed the Prompt-Aware Decoder (PAD), which guides the decoding process by generating task-driven signals from the interaction between the expression and visual features. Extensive experiments demonstrate the superiority of IPDN.

\section*{Acknowledgements}
This work was supported by National Key R\&D Program of China (No.2023YFB4502804), the National Science Fund for Distinguished Young Scholars (No.62025603), the National Natural Science Foundation of China (No. U22B2051, No. U21B2037, No. 62072389, No. 62302411, No. 624B2118), the Natural Science Foundation of Fujian Province of China (No.2021J06003), and China Postdoctoral Science Foundation (No. 2023M732948).

\bibliography{aaai25}

\end{document}